\documentclass[10pt,twocolumn,letterpaper]{article}

\usepackage[pagenumbers]{cvpr} 

\usepackage{lineno}
\usepackage{multirow}
\usepackage{svg}
\usepackage{algorithm}
\usepackage{algorithmic}
\usepackage{array}
\usepackage{booktabs} 
\definecolor{cvprblue}{rgb}{0.21,0.49,0.74}
\usepackage[pagebackref,breaklinks,colorlinks,allcolors=cvprblue]{hyperref}

\def\paperID{6707} 
\def\confName{CVPR}
\def\confYear{2025}

\title{AutoURDF: Unsupervised Robot Modeling from Point Cloud Frames 
\\ Using Cluster Registration}
\author{
Jiong Lin\quad Lechen Zhang\quad Kwansoo Lee\quad Jialong Ning\quad Judah Goldfeder\quad Hod Lipson \\
 Columbia University, Creative Machines Lab\\
 \tt\small \href{https://jl6017.github.io/AutoURDF/}{https://jl6017.github.io/AutoURDF/}
 }

\begin{document}
\maketitle
\begin{abstract}
Robot description models are essential for simulation and control, yet their creation often requires significant manual effort. To streamline this modeling process, we introduce AutoURDF, an unsupervised approach for constructing description files of unseen robots from point cloud frames. Our method leverages a cluster-based point cloud registration model that tracks the 6-DoF transformations of point clusters. Through analyzing cluster movements, we hierarchically address the following challenges: (1) moving part segmentation, (2) body topology inference, and (3) joint parameter estimation. The complete pipeline produces robot description files that are fully compatible with existing simulators. We validate our method across a variety of robots, using both synthetic and real-world scan data. Results indicate that our approach outperforms previous methods in registration and body topology estimation accuracy, offering a scalable solution for automated robot modeling. 
\end{abstract}
\section{Introduction}
\label{sec:intro}
\begin{figure}[!t]
  \centering
  \includegraphics[width=\columnwidth]{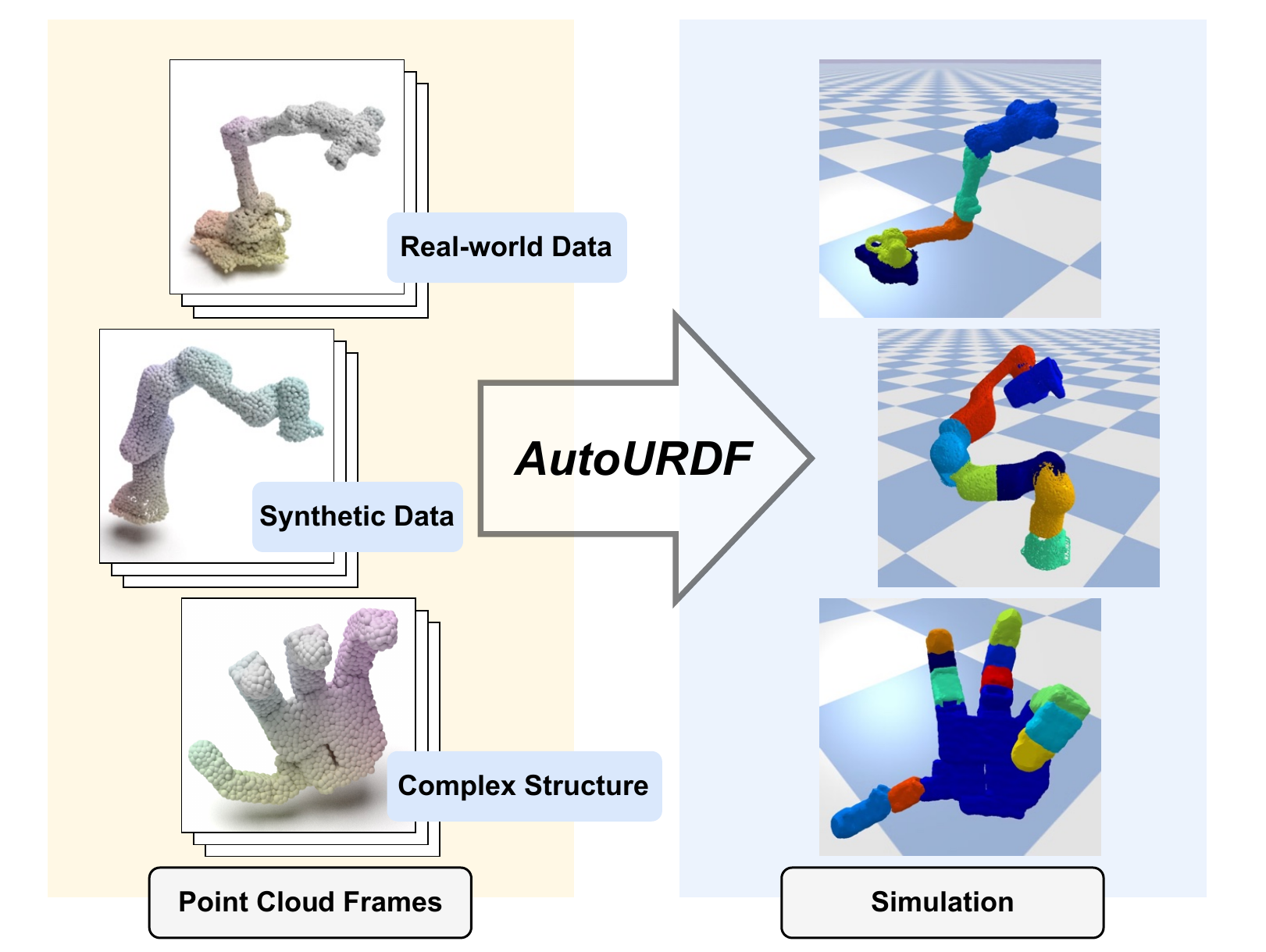}
  \caption{
    We present AutoURDF, a novel framework that derives robot description files from time-series point cloud frames.
    We validate our method across a diverse range of robots, including both synthetic and real-world scan data.
  }
  \label{fig_1}
\end{figure}
Accurate and structured representations of robots are essential for applications such as real-time control, motion planning, and physics-based simulation. Robot description files, such as the Unified Robot Description Format (URDF) \cite{urdf_roswiki}, are among the most widely used representations and explicitly capture robot geometry, kinematics, and dynamic properties.  Over time, new formats such as MJCF for MuJoCo \cite{todorov2012mujoco} and USD for Isaac Gym \cite{makoviychuk2021isaac} have been introduced to allow scene descriptions and parallel simulations. Despite these advancements, customizing basic robot models still requires manual effort, often involving CAD model conversions \cite{brawner_solidworks_urdf_exporter} or tedious XML file modifications. This challenge has driven robotics and computer vision researchers to explore data-driven methods to automate the robot modeling process.

\textit{Robot self-modeling} \cite{doi:10.1126/scirobotics.aau9354} has emerged as a vital approach in enabling robots to autonomously discover their own body kinematics. Previous works in this field utilize sensorimotor data, integrating depth images \cite{doi:10.1126/scirobotics.abn1944}, RGB images \cite{liu2024differentiablerobotrendering, hu2023teaching} or IMU sensors \cite{doi:10.1126/scirobotics.adh0972} with motor signals to achieve accurate self-modeling. Another related field, \textit{Articulated Object Modeling} \cite{li2020category}, aims to reconstruct the kinematic structure of articulated rigid bodies from visual data. Prior research has primarily concentrated on everyday objects \cite{10.1145/3272127.3275027, huang2021multibodysync,jiang2022ditto,mandi2024real2code}, such as scissors, laptops, and drawers, which typically have only a few moving parts and relatively simple kinematic structures. Robots, however, consist of serially connected joints and, in some cases, multibranched links. Additionally, large-scale datasets capturing diverse robot morphologies are still lacking. These factors make it challenging to effectively apply supervised learning methods designed for articulated objects to more complex robotic structures.

Here, we present AutoURDF, an unsupervised approach that constructs robot description files, specifically URDFs, from time-series point cloud frames. Rather than relying on sensorimotor data, our method derives the robot structure purely from visual data. Furthermore, it does not require ground-truth annotations or manual intervention, making it scalable to a wide range of robots. The key objective of this work is illustrated in Figure \ref{fig_1}.

Our method follows a hierarchical process, beginning with clustering the point cloud from the first frame based on point positions. Next, we compute a series of 6-DoF transformations across all time steps by registering the initial point clusters to each frame. From the motion patterns of these clusters, we segment them into distinct moving parts. Using the minimum spanning tree (MST) \cite{kruskal1956shortest} and segmented parts, we identify link connections and infer the body topology. Transformation matrices are then calculated for each child part relative to its parent, allowing us to derive the joint parameters. Finally, we generate the URDF file and validate the reconstructed robots in the PyBullet \cite{coumans2015bullet} simulator. Our contributions are as follows:
\begin{itemize}
  \item We propose a novel framework, AutoURDF, that builds robot description models purely from point cloud data.
  \item We design a neural network for cluster-based point cloud registration in complex articulated objects. 
  \item We implement a description file generation pipeline that outputs URDF files, including XML formatting and mesh file creation.
\end{itemize}
\section{Related Work}
\label{sec:related_work}
\begin{figure}[!t]
  \centering
  \includegraphics[width=\columnwidth]{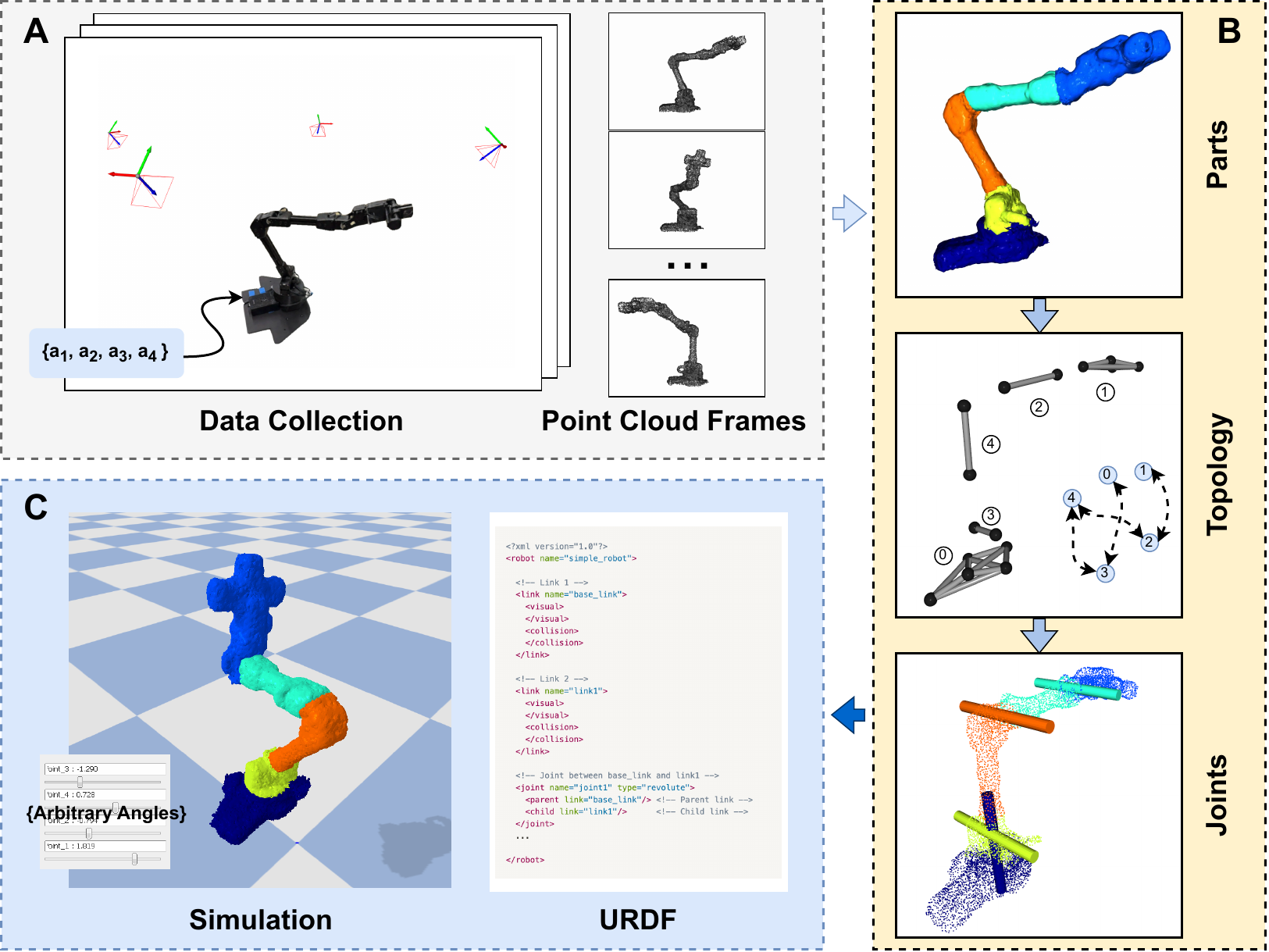}
  \caption{
    \textbf{AutoURDF overview.}
    Our method provides a complete pipeline for the automated construction of robot description files.
    \textbf{(A). Data Collection:} By commanding robots with randomly sampled motor angle sequences, we capture the corresponding time-series point cloud frames.
    \textbf{(B). Three Substeps:} We tackle the problem in three substeps: 1) part segmentation, 2) body topology inference, and 3) joint parameter estimation.
    \textbf{(C). Description File Generation:} The final output is a URDF file that defines the robot's links, joints, and collision properties. 
    We successfully build and simulate the description model for a WX200 robot arm from real-world scan data.
  }
  \label{fig_2}
\end{figure}

\begin{figure*}[!t]
  \centering
  \includegraphics[width=\textwidth]{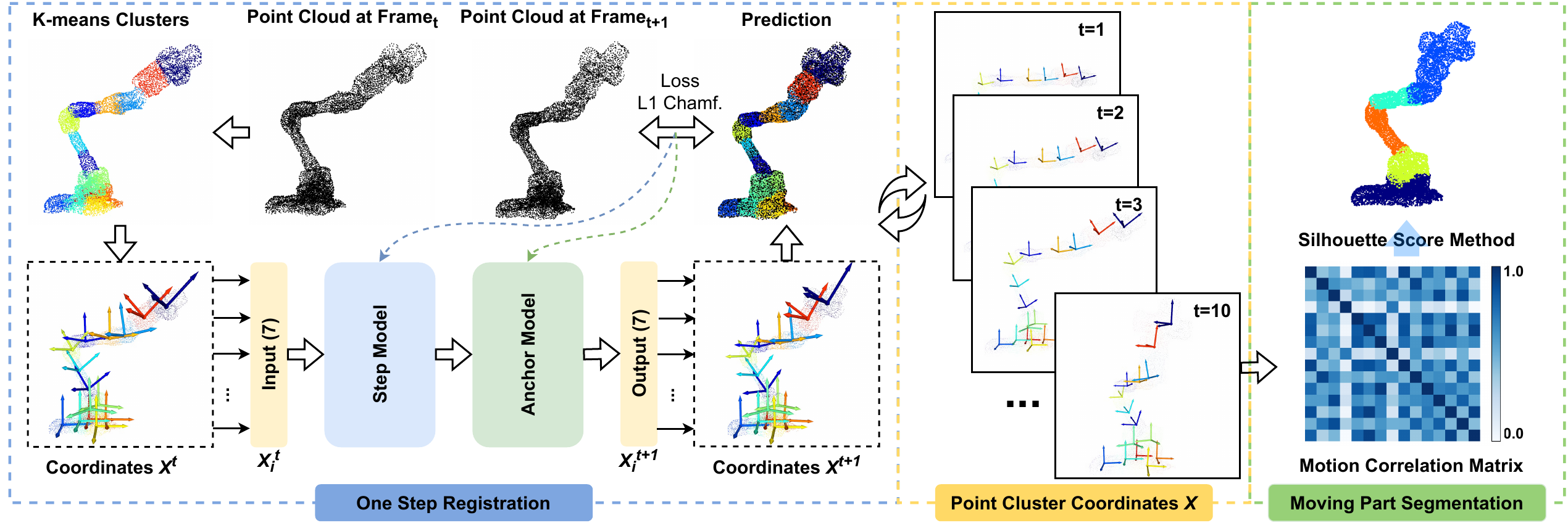}
  \caption{
    \textbf{Point Cluster Registration and Part Segmentation.}
    We tackle the part segmentation problem by extracting 6-DoF transformations from the registration of a set of point clusters. 
    For each pair of consecutive frames, we begin by initializing point clusters using the K-means clustering algorithm. Each point cluster is assigned a 6-DoF coordinate, ${X^t_i}$, predicted from the previous registration step.  
    Next, we employ a shared neural network as the Step Model to predict the target frame's point cluster coordinates, ${X^{t+1}_i}$. Following the rotation representation study \cite{geist2024learning,bregier2021deep,zhou2019continuity}, we represent it as a 7-dimensional vector, combining Cartesian coordinates and quaternion orientation.
    The predicted coordinates are passed through an Anchor Model that computes cluster transformations, starting from the initial coordinates ${X^1}$.
    This process is repeated across frames in the sequence, tracing the motion of a set of 6-DoF coordinate frames.
    Finally, based on the time-series 6-DoF transformations, we compute a correlation matrix between the point clusters to segment distinct moving parts.
  }
  \label{fig_3}
\end{figure*}

{\bf{Robot self-modeling.}} 
Task-agnostic self-modeling \cite{doi:10.1126/scirobotics.aau9354} enables robots to autonomously construct and update their own kinematic models from interaction data, without requiring specific task information. Recent advances in neural implicit representations have shown promising results in reconstructing dynamic 3D shapes \cite{pumarola2021d, luiten2023dynamic}. By applying spatial implicit functions \cite{park2019deepsdf, mildenhall2021nerf, kerbl3Dgaussians} to robot modeling, Chen et al. \cite{doi:10.1126/scirobotics.abn1944} demonstrated effective shape prediction by modeling robot morphology as a spatial query model conditioned on motor commands. Liu et al. \cite{liu2024differentiablerobotrendering} further extended this approach with a fully differentiable robot body rendering model. However, both methods require robot control information for model supervision, limiting their autonomy. Additionally, implicit function-based robot models are powerful for visual supervision and shape prediction, but they are not directly compatible with physics-based simulators, and accurately and efficiently calculating physical interactions with them remains challenging. On the other hand, Ledezma et al. \cite{doi:10.1126/scirobotics.adh0972} take a data-driven approach to robot modeling, explicitly representing the robot's morphology through topology and kinematic parameters. However, their method requires IMU sensors on each moving part, which greatly increases setup complexity as the robot’s degrees of freedom (DoF) increase. Our method is motivated by similar goals to task-agnostic self-modeling approaches, aiming to enable autonomous robot modeling without specific task constraints. Unlike these approaches, our method relies solely on raw point cloud frames, without needing forward kinematics, degrees of freedom (DoF), or motor control data. Additionally, our approach maintains an explicit representation of the robot's kinematics and geometry, ensuring direct compatibility with robot simulators.\\

\noindent{\bf{Articulated objects modeling.}} 
\textit{PartNet-Mobility} dataset \cite{Mo_2019_CVPR} has advanced research in supervised learning for articulated objects modeling.
Most existing studies focus on everyday objects with only one or a few degrees of freedom \cite{jiang2022ditto, jiayi2023paris, mu2021sdf, heppert2023carto, zhang2023flowbot++, jain2021screwnet, song2024reacto, wang2019shape2motion, Chu_2023_CVPR,nie2022structure}.
Few works \cite{liu2023building, noguchi2022watch, mandi2024real2code, xu2020rignet, chen2024urdformer} have explored the challenging task of reconstructing complex kinematic structures with more than 10 degrees of freedom.
Among these, \textit{Real2Code} \cite{mandi2024real2code} and \textit{URDFormer} \cite{chen2024urdformer} are recent work that reconstructs the kinematic structure of actuated objects through code generation.
However, these works assume moving parts are connected to a single parent part, which is not applicable to robotic structures with a series of joints for each kinematic branch.
\textit{Watch-It-Move} \cite{noguchi2022watch} and \textit{Reart} \cite{liu2023building} are two notable works that reconstruct the kinematic structure for multiple kinds of robots.
However, these methods use custom interaction code: \textit{Watch-It-Move} repositions the output robot models by applying rototranslation to each joint, and \textit{Reart} reanimates the model by specifying target points for each part.
In contrast to the aforementioned methods, our approach does not rely on ground-truth annotations for training supervision and 
generates standard URDF files that integrate seamlessly with kinematics computation in simulators.

To the best of our knowledge, our method is the first to construct functional description files for complex robotic structures (with up to 18 degrees of freedom and serially connected links) from unlabeled point cloud data.
\section{Methodology}
\label{sec:method}
Our goal is to reconstruct a robot description file from a time series of point cloud frames by extracting the necessary structural and motion information.
This process involves segmenting the robot links, identifying the connections between them, determining joint parameters,
and capturing the geometry of each link as mesh files. In this paper, we focus on revolute joints, the most common type in robotic structures, and assume a tree-like structure without closed kinematic loops.

\subsection{Problem Formulation and Method Overview}
For each robot reconstruction task, we collected a time series of whole-body point cloud frames,
\(\boldsymbol{\mathcal{P}} = \{ P^t \in \mathbb{R}^{3 \times N} \}_{t \in [1, T]}\), 
where \(P^t\) is the point cloud at time \(t\) and \(N\) is the number of points in each frame. Our goal is to extract three components:

\begin{itemize}
    \item \textbf{Segmented parts}: denoted as link point clouds \(\boldsymbol{\mathcal{L}} = \{ L_i \in \mathbb{R}^{3 \times N_i} \}_{i \in [1, M]}\), where \(L_i\) is the point cloud of the \(i\)-th link, and \(M\) is the total number of links.

    \item \textbf{Body topology}: represented as graph \(\boldsymbol{\mathcal{G}} = (I, E)\), where \(I = \{ i \in [1, M] \}\) is the set of link indices and \(E = \{ (i_p, i_c) \}_{i_p, i_c \in I, i_p \neq i_c}\) is the set of edges connecting the parent and child links.

    \item \textbf{Joint parameters}: represented as \(\boldsymbol{\mathcal{J}} = \{ J_i\}_{i \in [1, K]}\), where \(J_i\) contains the 6-DoF parameters of the \(i\)-th joint, and \(K\) is the number of joints. For the tree-like structure, we have \(K = M - 1\).
\end{itemize}
To achieve this goal, we design a rigid body registration algorithm to track the motion of a set of point clusters, initialized at time step \( t=1 \). We define the set of point clusters as \( \boldsymbol{\mathcal{C}} = \{ C^t_i \in \mathbb{R}^{3 \times N_i} \}_{t \in [1, T], i \in [1, S]} \), where \( S \) is a hyperparameter that determines the number of clusters.
The motion of the clusters is represented by a set of position and orientation coordinates, \( \boldsymbol{\mathcal{X}} = \{ X^t_i = (x^t_i, q^t_i) \}_{t \in [1, T], i \in [1, S]} \), where \( x \in \mathbb{R}^{3} \) represents the Cartesian center of the point cluster, and \( q \in \mathbb{R}^{4} \) is the quaternion orientation of the cluster. These coordinates are used to compute a correlation matrix between point clusters.
Based on this correlation matrix, \( \boldsymbol{\mathcal{M}}^{S \times S} = \{\rho(X_i, X_j)\}_{i, j \in [1, S]} \), we segment the clusters into distinct moving parts, denoted as the predicted links \( \boldsymbol{\mathcal{\hat{L}}} \). We then apply the MST algorithm on the position components of \(\boldsymbol{\mathcal{X}}\) to infer the body topology \(\boldsymbol{\mathcal{\hat{G}}}\).
Finally, we estimate the joint parameters \(\boldsymbol{\mathcal{\hat{J}}}\) by computing the homogeneous transformation matrices for pairs of clusters associated with connected links.

In summary, we derive $\boldsymbol{\mathcal{\hat{L}}},\boldsymbol{\mathcal{\hat{G}}},\boldsymbol{\mathcal{\hat{J}}}$ from 
$\boldsymbol{\mathcal{P}}$ by solving for transformation coordinates $\boldsymbol{\mathcal{X}}$ and correlation matrix $\boldsymbol{\mathcal{M}}$.

\subsection{Cluster Registration and Part Segmentation}
\label{subsec:L}
Traditional point cloud registration algorithms \cite{121791} typically focus on rigid body alignment, while non-rigid registration methods handle more complex deformation scenarios, such as those involving the human body, cloth, or soft objects. In these cases, 3D coordinates (e.g., x, y, z) and graph representations are often used to model deformation. 
Robots, however, are articulated rigid bodies with 1-DoF joints, which can be viewed as a simplified case of non-rigid structures.
We leverage the coherent motion of rigid parts to extract accurate 6-DoF transformations for each point cluster.

As shown in Figure \ref{fig_3}, 
The \textbf{Step Model} registers point clusters from time step $t$ to the ground truth at $t+1$, while the \textbf{Anchor Model} registers clusters from the first time step directly to the ground truth at $t+1$. The initial coordinates represent the 6-DoF poses (translation and rotation) of the initially segmented clusters, serving as the starting point for registration.
Inspired by PointNet \cite{qi2017pointnet, qi2017pointnet++}, we designed a new lightweight neural network that computes coordinates using a shared MLP module, leveraging the permutation-invariant property to ensure that the input cluster order does not affect the output. Details on the model architecture and 6-DoF pose comparisons are provided in the appendix B.

The loss function Eq. \ref{eq1} is a bi-directional L1 Chamfer distance \cite{borgefors1988hierarchical, ravi2020pytorch3d} between the predicted point clusters and the target whole-body point cloud.
This loss function serves to optimize the network parameters, which in turn updates the output coordinates.
\begin{equation}
  \label{eq1}
  \begin{aligned}
  \mathcal{L}_{\text{Chamfer}}^1(\hat{P}^t, P^t) = 
  &\sum_{x=1}^{N} \min_{y} \left\| \hat{P}_x^t - P_y^t \right\|_1 \\
  +&\sum_{y=1}^{N} \min_{x} \left\| \hat{P}_x^t - P_y^t \right\|_1
  \end{aligned}
\end{equation}
$\hat{P}^t =\bigcup_{i=1}^{S} {}^{w}\hat{C}_i^t$ is the predicted point cloud for the entire body, 
integrated from the set of point clusters in the \textit{world coordinate frame} at time step $t$, $\{ {}^{w}\hat{C}^t_i \}_{i \in [1, S]}$.
$P^t$ is the target whole body point cloud at the same time step.
Step registration includes a resampling module, where the predicted coordinates $X^{t+1}$  are used as the center for clustering the next frame's point cloud $P^{t+1}$. Consequently, the step matching model performs registration over one frame interval.
The Anchor Model takes as input the predicted coordinates from the Step Model and the initial point cluster from the first frame. It refines the coordinates to capture the transformation from the initial step to the target frame.

We demonstrate our registration pipeline, whose input is a whole body point cloud video $\boldsymbol{\mathcal{P}}$ and output are the clustered point cloud set $\boldsymbol{\mathcal{C}}$ and the correlated 6D coordinates $\boldsymbol{\mathcal{X}}$, 
in algorithm \ref{alg:alg1}.
In the clustering and Chamfer distance calculations, we use point clusters in the world coordinate frame. For optimization in the Step Model $\boldsymbol{\mathcal{F}_S}$ and Anchor Model $\boldsymbol{\mathcal{F}_A}$, we work with point clusters in the local coordinate system, represented by the tracked 6-DoF coordinates associated with each cluster.
We employ two clustering algorithms: K-means\cite{macqueen1967some} in the resampling module, where cluster centers are provided as input, and K-means++\cite{arthur2006k} in the initialization step, where the hyperparameter $S$ (the number of clusters) is specified.
$H$ represents the set of homogeneous transformations in SE(3), and $\theta$ denotes the parameters of the neural network.
\begin{algorithm}[H]
\caption{Point Cluster Registration}\label{alg:alg1}
\begin{algorithmic}
  \STATE \textbf{Input}: Point Cloud Frames $\boldsymbol{\mathcal{P}}$\par
  \STATE \textbf{Output}: Point Clusters $\boldsymbol{\mathcal{C}}$, Coordinates $\boldsymbol{\mathcal{X}}$ \par
  \STATE \textbf{Initialize}: at step $1$, ${}^w C^1 = \text{Clustering}_1(P^1, N_{seg}=S)$ \par
  \STATE ${X}^1 = \{X^1_i = (x=mean(C^1_i), \alpha=zeros(3))\}_{i \in [1, S]}$ \par
  \STATE Add $ X^1$, $ {}^l C^1$ to $\boldsymbol{\mathcal{X}}$, $\boldsymbol{\mathcal{C}}$ (${}^lC$ in local coordinate frame)\par 
\textbf{for} t = 1, ..., T \textbf{do} \par
\hspace{1em} Load $X^t, {}^l C^t$ from $\boldsymbol{\mathcal{X}}$, $\boldsymbol{\mathcal{C}}$ \par
\hspace{1em} \textbf{Step Registration: }\par
\hspace{2em} ${}^w \hat{H}_l^{t+1} \leftarrow \hat{X}^{t+1} = \boldsymbol{\mathcal{F}_S}(X^{t}, {}^l C^t, \theta_S)$\par
\hspace{2em} where ${}^w H_l=\{{}^w H_{l,i} \in SE(3)\}_{i \in [1, S]}$\par
\hspace{2em} ${}^w\hat{C}^{t+1} \leftarrow \{{}^w \hat{H}_{l,i}^{t+1} \cdot {}^l C_i^{t} \}_{i \in [1, S]}  $\par
\hspace{2em} $\hat{P}^{t+1} =\bigcup_{i=1}^{S} {}^w\hat{C}_i^{t+1}$ \par
\hspace{2em} Find $\theta_S^* = \arg \min_{\theta} \mathcal{L}^1_{Chamfer}(\hat{P}^{t+1}, P^{t+1})$ \par
\hspace{2em} Update $\hat{X}^{t+1} $, $\hat{C}^{t+1}$ with $\theta^*_S $\par
\hspace{2em} \textit{Resample}: \par
\hspace{2em} ${}^w\hat{C}^{t+1} =\text{Clustering}_2(P^{t+1}, \text{Centers}=\hat{x}^{t+1})$ \par
\hspace{2em} where the $\hat{x}$  is the position term of $\hat{X}$ \par
\hspace{2em} Transfer ${}^w\hat{C} \rightarrow {}^l\hat{C} $  \par
\hspace{1em} \textbf{Anchor Alignment: }\par
\hspace{2em} $\bar{X}^{t+1} = \boldsymbol{\mathcal{F}_A}(\hat{X}^{t+1}, {}^lC^1, P^{t+1})$ \par
\hspace{1em} Add $\bar{X}^{t+1}$, ${}^l\hat{C}^{t+1}$ to $\boldsymbol{\mathcal{X}}$, $\boldsymbol{\mathcal{C}}$ \par

\textbf{end for} \par
\end{algorithmic}
\label{alg1}
\end{algorithm}
As shown in equation \ref{eq2} and equation \ref{eq3}, we compute the correlation matrix $\boldsymbol{\mathcal{M}}$ between the point clusters based on the transformation coordinates $\boldsymbol{\mathcal{X}}$. The distance is calculated as the Euclidean distance between positions and the geodesic distance between orientations. Here, $\alpha$ is a scaling parameter automatically calculated from the bounding box of the point cloud. The result is normalized to a range between 0 and 1 to facilitate alignment across multiple sequences.
\begin{equation}
  \label{eq2}
  \rho (X_i, X_j)_{i, j \in [1,S]} = \frac{\sum_{t=1}^{T} \boldsymbol{\mathcal{D}}(X_i^t, X_j^t )}
  {\max_{i,j} \sum_{t=1}^{T}\boldsymbol{\mathcal{D}}(X_i^t, X_j^t )}
\end{equation}
\begin{equation}
  \label{eq3}
  \boldsymbol{\mathcal{D}}(X_i^t, X_j^t ) = \alpha \cdot \textbf{d}_{Euc}(x_i^t, x_j^t) + \textbf{d}_{Geo}(q_i^t, q_j^t)
\end{equation}
Clusters with highly correlated coordinate sequences (indicating minimal pose differences) are grouped into the same moving parts. We then merge the point clusters within each moving part to form the predicted links $\boldsymbol{\mathcal{\hat{L}}}$, which represent sets of point clouds that capture the shape of the robot links.

We use the Silhouette Score method \cite{rousseeuw1987silhouettes} to determine the optimal number of segments based on the Motion Correlation Matrix. 
\subsection{Topology Inference}
\label{subsec:G}
\begin{figure}[!t]
  \centering
  \includegraphics[width=\columnwidth]{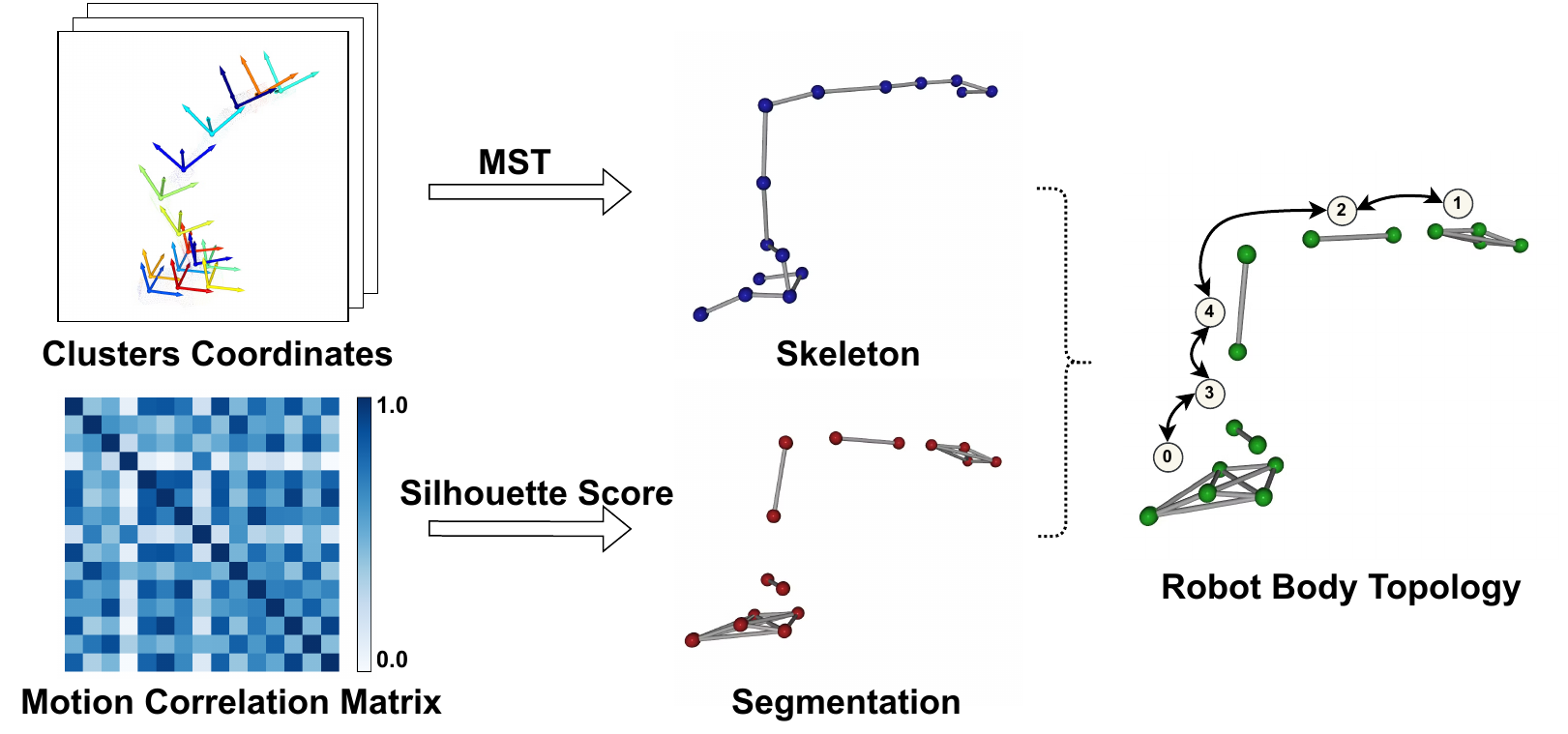}
  \caption{
    \textbf{Topology Inference.}
    We infer the robot's body topology by merging the result of the MST algorithm on position terms $x$ and the segmented point clusters. In the segmentation graph, the nodes represent the center positions of the clusters; the edges indicate the motion correlated between the connected nodes.
  }
  \label{fig_4}
\end{figure}
We infer the body topology using two graphs: (1) the graph of segmented point clusters, where nodes represent cluster indices and edges represent inter-cluster connections; and (2) the minimum spanning tree constructed using the accumulated positional distance between point clusters as edge weights.
We demonstrate the process in Figure \ref{fig_4}, and detailed algorithm in Appendix B.
The MST algorithm is applied to the point clusters, where the weight of each edge is determined by the accumulated position distance between the point clusters.
We identify edges in the MST that connect point clusters from different moving parts, which correspond to the connections between robot links.
The resulting graph $\boldsymbol{\mathcal{\hat{G}}}$ represents the body topology of the robot, with the nodes as the link index and the edges as the connections between the links.
To comply with the URDF format, we assign the robot's base link as the root node. We rank all links by their total pose variation across frames and select the one with the least variation as the base link.
Starting from this base link, we traverse the graph as a directed tree to establish the parent-child relationships between the links.

\subsection{Joint Estimation}
\label{subsec:J}
Referring to the merged point clusters, we compute the average 6D coordinates of each link
.
In joint estimation, we convert the pose coordinates \(X = (x, q)\) into a homogeneous transformation matrix \(H \in \mathrm{SE}(3)\). The base link is assigned the identity matrix \(I\), and each link's pose in the world frame is denoted by \(H_i\). The relative transformation from the parent link to the \(i\)-th link is then computed as $H_i^p = H_i \cdot H_p^{-1}$ where \(H_p\) is the pose of the parent link.

According to the body topology graph, we locate every parent-child pair of links and compute the time-series transformation matrices of each child link with respect to the parent link’s coordinate system.
For each 1-DoF revolute joint connecting a pair of links, the relative transformation yields a rotation matrix
$R \in SO(3)$ and a static translation vector $t \in \mathbb{R}^{3}$.
We extract the joint parameters $\boldsymbol{\mathcal{\hat{J}}}$ as the rotation axis and joint center, derived from the rotation matrix and translation vector, respectively.
Our method constrains SE(3) transformations to 1-DoF joint motions using Euler’s rotation theorem, parameterizing motion with a fixed point,
rotation axis, and angle, making it naturally compatible with URDF.

\subsection{URDF Generation}
\begin{figure}[!t]
  \centering
  \includegraphics[width=\columnwidth]{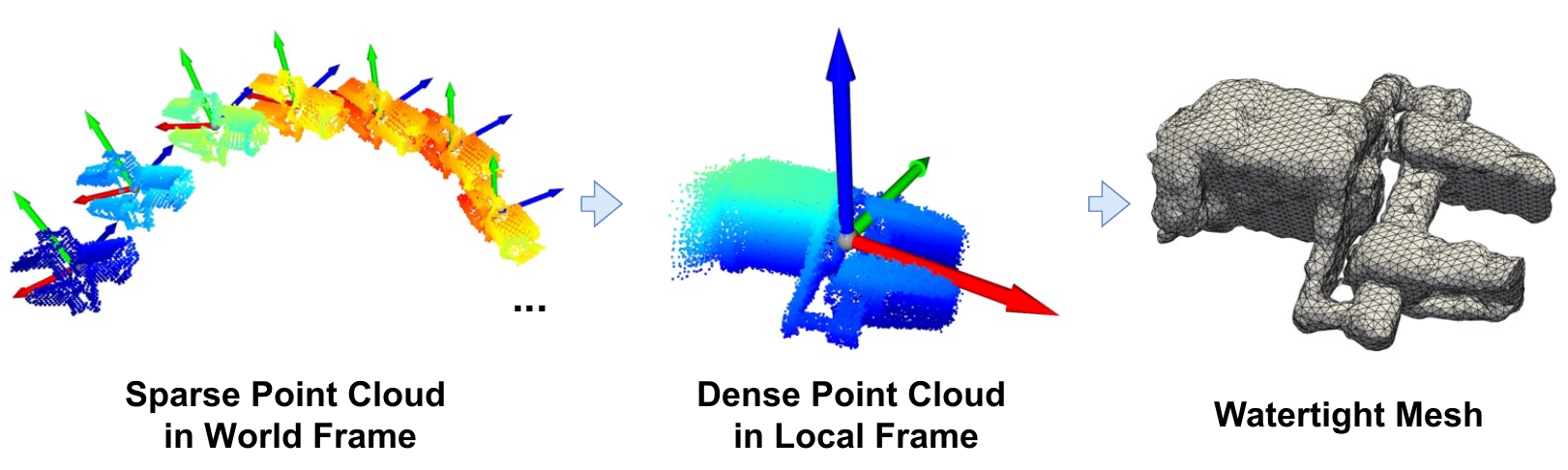}
  \caption{
    \textbf{Point Cloud to Mesh.} 
    For each segmented link, we integrate the sparse point cloud data from the world coordinate system at each time step to form a dense point cloud in the local frame. 
    The dense point cloud is constructed by combining 10 frames of data. 
    We then apply the marching cubes\cite{lorensen1987marching} algorithm to convert this dense point cloud into a mesh file. 
    The resulting mesh files are high-quality and watertight, as demonstrated by the example of the WX200 robot arm's end-effector shown in the figure.
    }
  \label{fig_5}
\end{figure}

The URDF contains an XML file that defines the properties of the robot’s links and joints, along with a set of mesh files representing the geometric details of each link. Based on the predicted body topology and joint parameters, we generate the XML file specifying the links and joints of the robot. 
The pose coordinates of each link, which represent the transformation from the link’s local coordinate system to the world coordinate system, allow us to express all link point clouds in a consistent world frame. This enables integration of sparse point clouds across frames into a dense point cloud in the local frame.
As shown in Figure \ref{fig_5}, we combine data from 10 frames to construct a dense point cloud, which we then process using the marching cubes algorithm\cite{lorensen1987marching} to convert the dense representation into a high-quality, watertight mesh. The resulting mesh files are used to define the geometric properties of each link in the URDF file. In this paper, we focus on kinematics and morphology, leaving the modeling of dynamic properties for future work.

In summary, we demonstrate the derivation of $\boldsymbol{\mathcal{\hat{L}}},\boldsymbol{\mathcal{\hat{G}}},\boldsymbol{\mathcal{\hat{J}}}$ in Section \ref{subsec:L}, \ref{subsec:G}, and \ref{subsec:J}, which together enable URDF generation.
\section{Experiments}
\label{sec:experiments}
\begin{figure}[!t]
  \centering
  \includegraphics[width=\columnwidth]{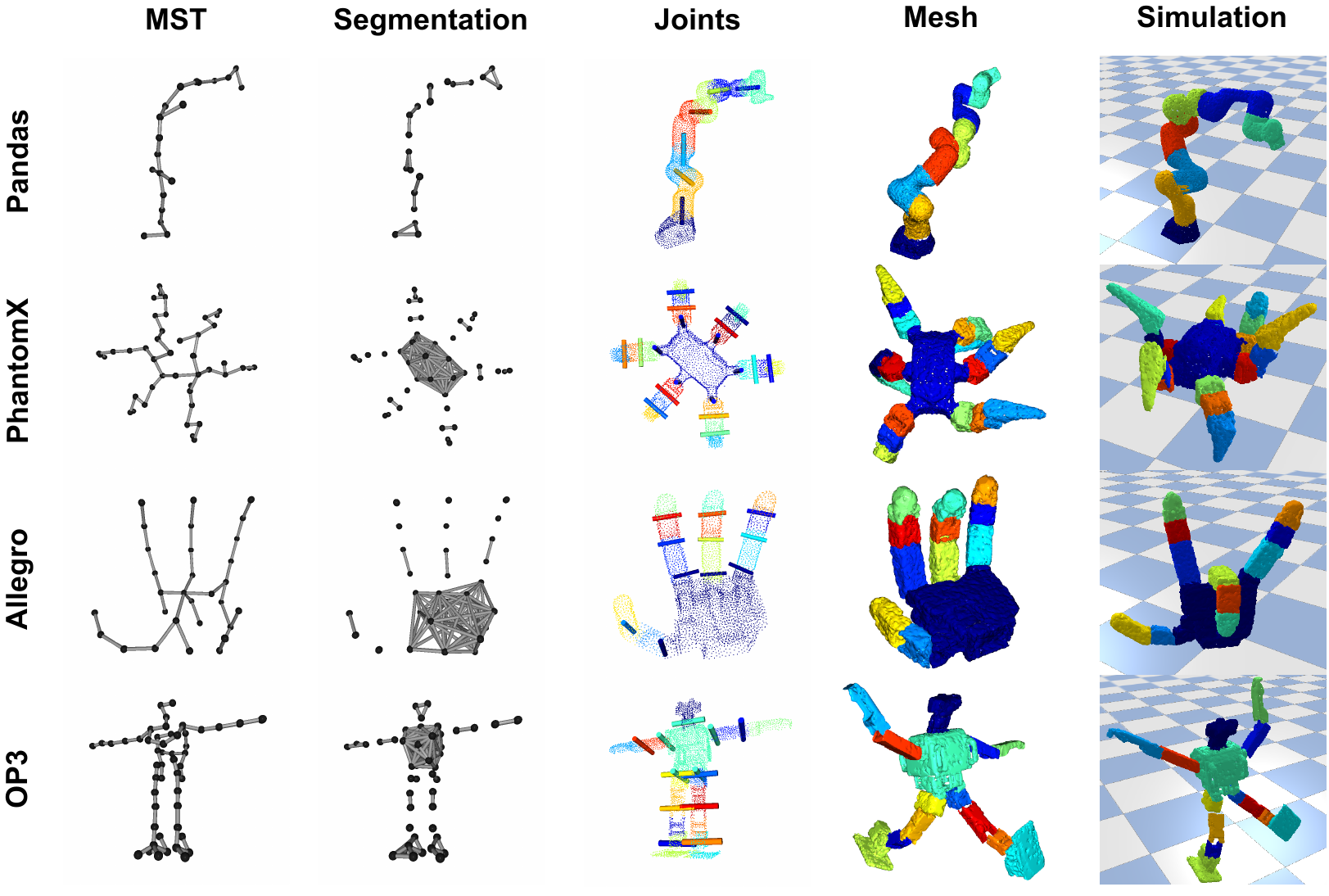}
  \caption{
    \textbf{Qualitative Results.}
    We illustrate the core stages of our method for various robots, including the minimum spanning tree, part segmentation, joint estimation results, mesh, and constructed URDFs for simulation. The colors of the joints indicate the parent link they are connected to. 
  }
  \label{fig_6}
\end{figure}



\begin{figure}[!t]
  \centering
  \includegraphics[width=\columnwidth]{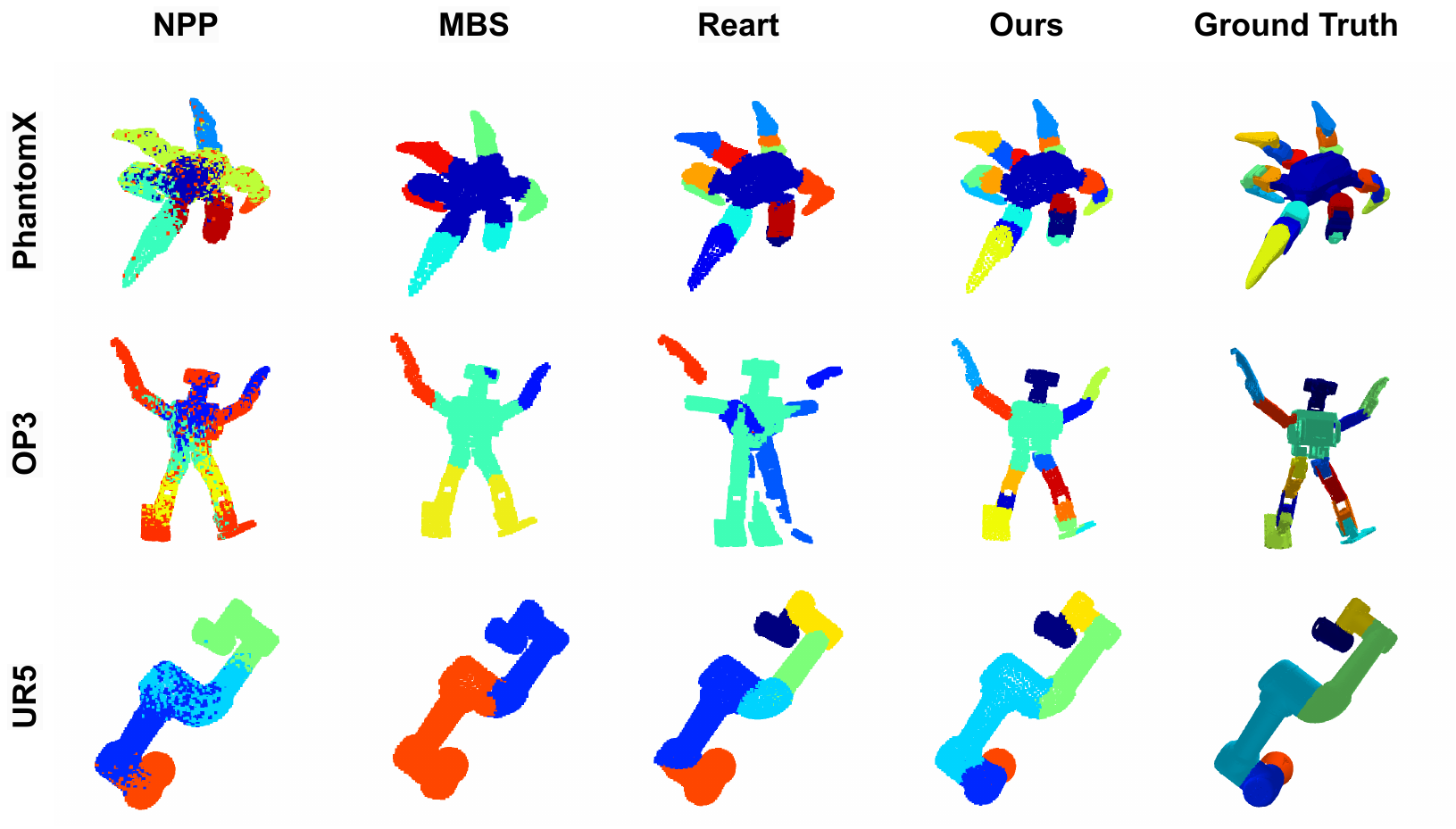}
  \caption{
    \textbf{Qualitative Comparison with Baseline Methods.}
    Given the same input sequence (10 point cloud frames), we compare our method with NPP \cite{hayden2020nonparametric}, MBS
    \cite{huang2021multibodysync}, and Reart \cite{liu2023building} for part segmentation and point cloud registration on the final frame.
  }
  \label{fig_7}
\end{figure}

\begin{figure}[!t]
  \centering
  \includegraphics[width=\columnwidth]{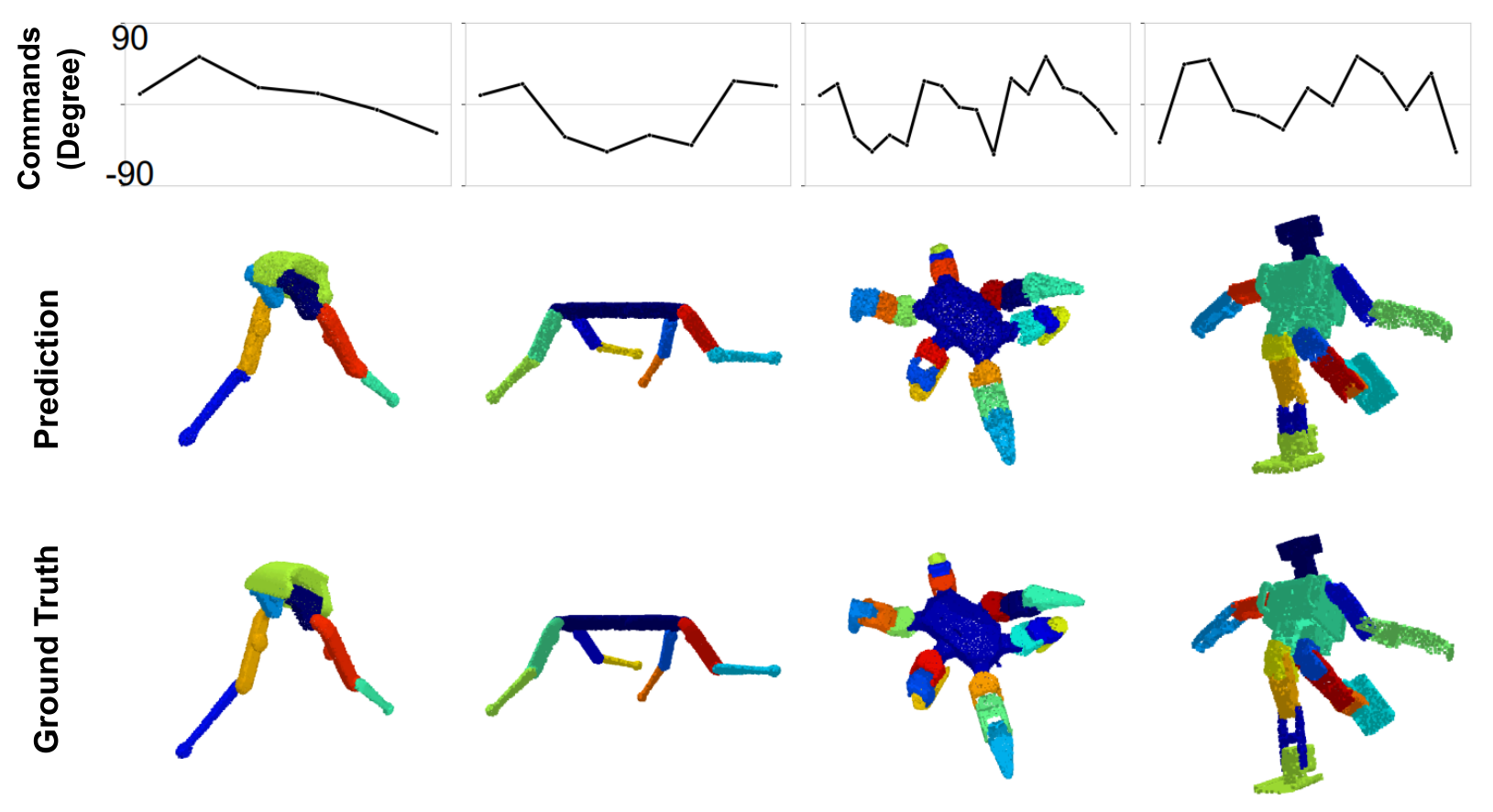}
  \caption{
    \textbf{Qualitative Results for Simulating New Configurations.}
    We compare the predicted URDFs with ground truth by applying the same motor commands. AutoURDF generates accurate description files, producing control results that closely match those of the ground-truth models. For visualization, we use the same colormap for both the predicted and ground-truth URDFs.
  }
  \label{fig_8}
\end{figure}

\subsection{Experiments Setup}
\label{subsect:e1}
{\bf{Dataset:}}
The validation dataset includes a variety of robots, including single-branch link robot arms (WX200, Franka Panda, UR5e) and multi-branch link robots (Bolt, Solo8, PhantomX, Allegro Hand, OP3 Humanoid), with degrees of freedom ranging from 5 to 18.
We evaluate our method on one point cloud video of a real-world scan of the WX200 robot arm and five point cloud videos of eight robots in simulated settings. Each video consists of 10 frames of point cloud data, down-sampled to 5,000 points per frame. For data collection, we fixed certain joints on the Allegro Hand and OP3 Humanoid and removed the end-effectors for the Franka Panda and UR5e. Further details on the data collection pipeline are provided in the Appendix A.

\noindent{\bf{Metrics:}}
We evaluate our method across four metrics: registration, body topology inference, joint estimation, and repose. \textbf{Registration} is measured using the average L1 Chamfer Distance (in millimeters), denoted as \( CD \) in our tables, which quantifies the alignment between the transformed whole-body point cloud from the first frame and each subsequent frame. \textbf{Body topology inference} is evaluated by comparing the predicted kinematic tree with the ground-truth kinematic tree using the tree editing distance \cite{pawlik2016tree, pawlik2015efficient}, denoted as \( TED \), to assess the accuracy of the inferred body topology.
We measure \textbf{joint estimation} by calculating the average angular difference between joints (in degrees) and the average normal distance between the predicted and ground-truth joint axes (in millimeters). We denote these two metrics as $E_{JD}$ and $E_{JA}$, unit in millimeters and degrees, respectively.
\textbf{Repose} accuracy, denoted as $CD_r$, is evaluated by commanding two URDFs with random new motor angles and computing the surface point cloud L1 Chamfer distance. It is important to note that Chamfer distance depends on the robot's scale and the density of the point cloud. In this work, we calculate the L1 Chamfer distance using the original scale of the robots, with 5,000 points sampled for alignment in all experiments.

\subsection{Qualitative Experiments}
\label{subsect:e2}
\begin{figure}[!t]
  \centering
  \includegraphics[width=\columnwidth]{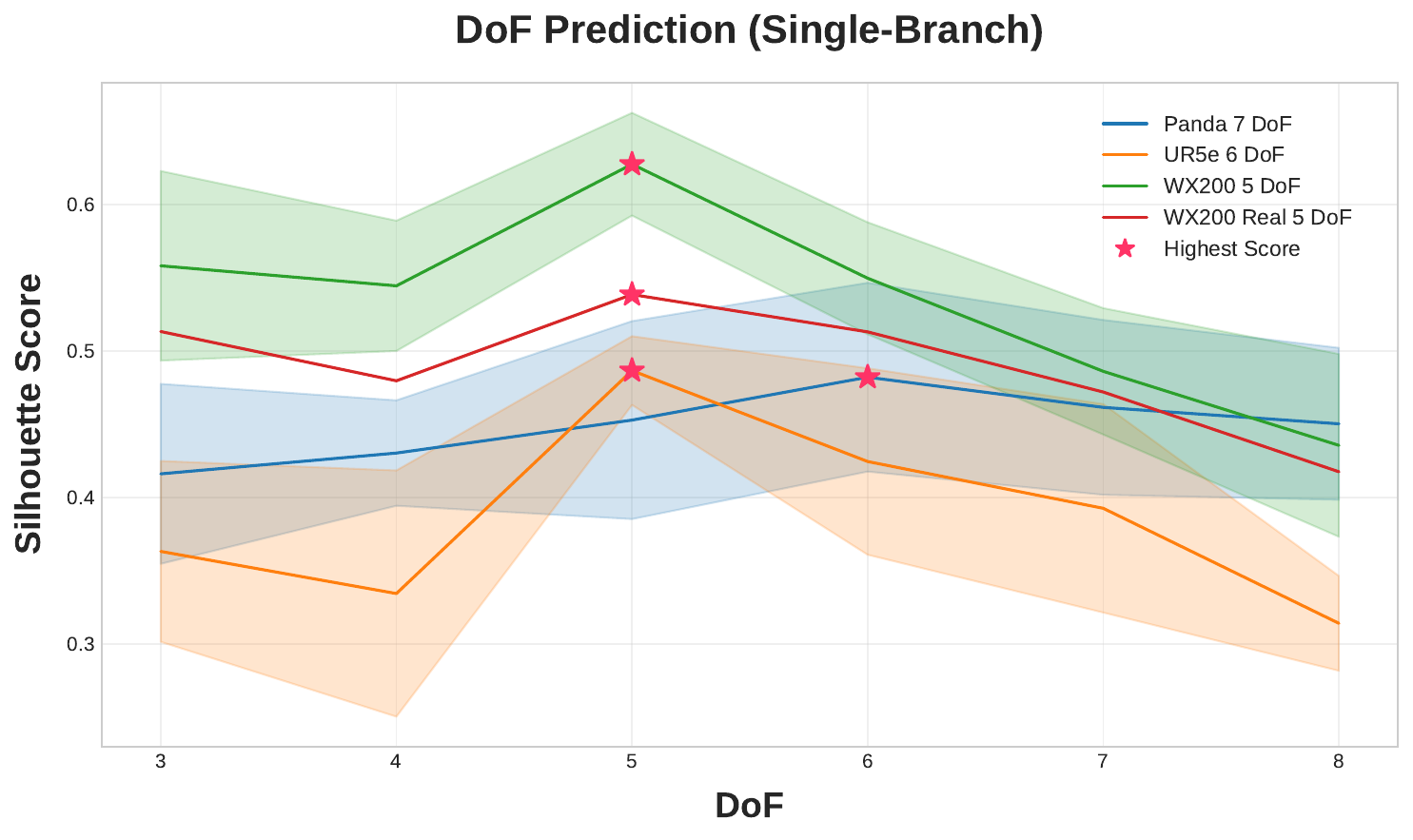}
  \includegraphics[width=\columnwidth]{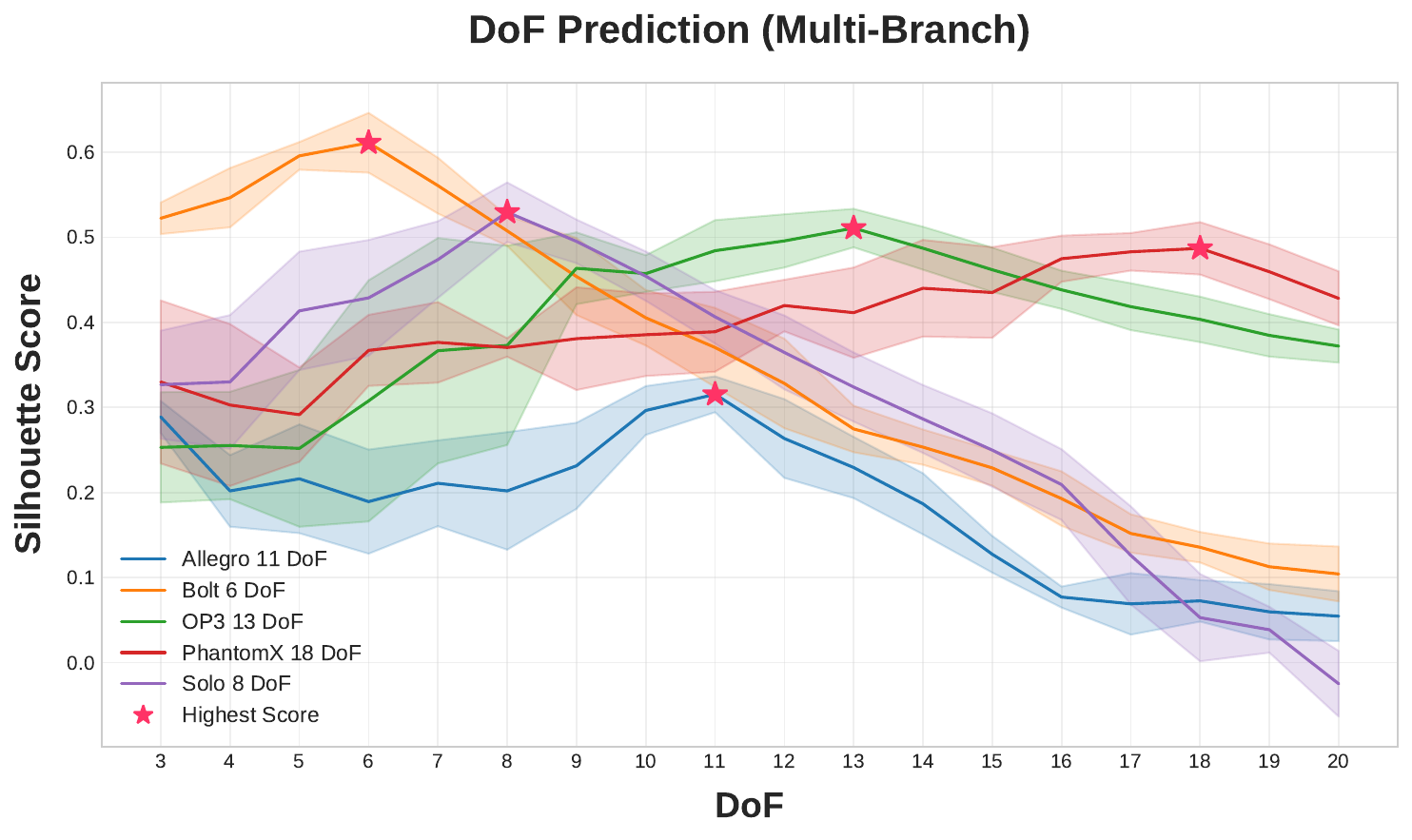}
  \caption{
    \textbf{DoF Prediction.} We plot the Silhouette Score for eight different robots across five point cloud sequences. The highest score indicates the optimal number of distinct moving parts, and thus the number of DoF. By averaging the Silhouette Scores across sequences, our method accurately predicts the correct number of DoF for all eight robots.
    }
  \label{fig_9}
\end{figure}

\begin{figure}[!t]
  \centering
  \includegraphics[width=\columnwidth]{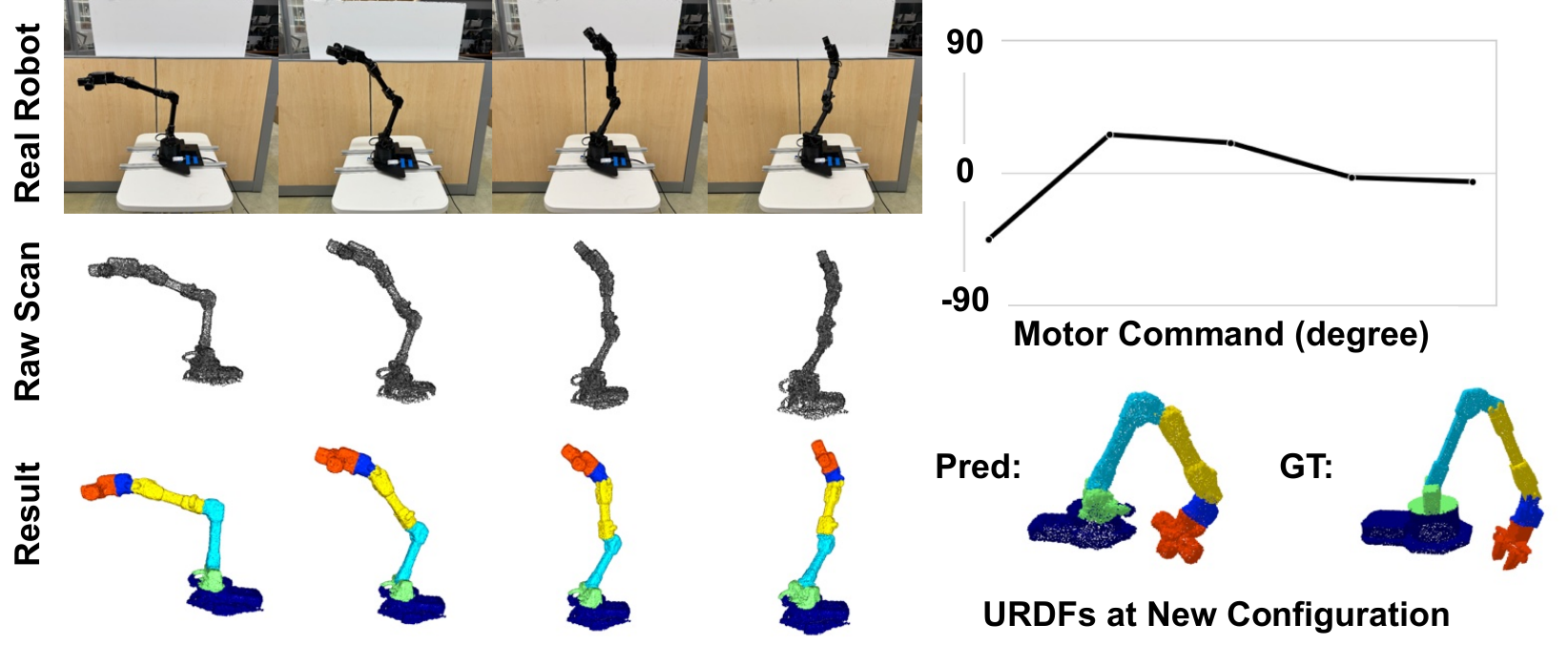}
  \caption{
    \textbf{Real World Experiment.} 
    The first row shows the WX200 robot arm at four different time steps under random motion commands. The second row displays the corresponding whole-body point cloud frames. The third row presents alignment results with meshes. 
    The right column compares the predicted URDF with the ground-truth model under a new configuration.
    }
  \label{fig_10}
\end{figure}
\begin{figure}[!t]
  \centering
  \includegraphics[width=\columnwidth]{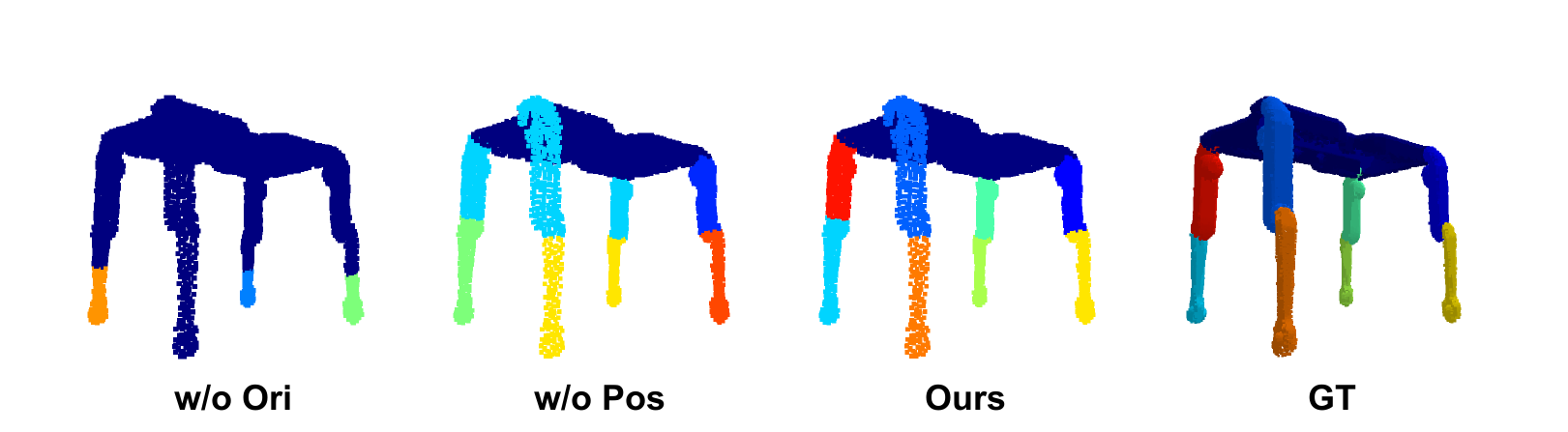}
  \caption{
  \textbf{Ablation study on Correlation Matrix.}
  We compare the distance equation \ref{eq3} under three conditions: without the orientation term $\textbf{d}_{Geo}$, without the position term $\textbf{d}_{Euc}$, and with the complete calculation.
    }
    \label{fig_11}
\end{figure}

\begin{table*}[!t]
\centering
\captionsetup{width=\textwidth}
\resizebox{\textwidth}{!}{
\renewcommand{\arraystretch}{1.2}
\begin{tabular}{
  p{0.06\textwidth}|
  p{0.11\textwidth}|
  >{\centering\arraybackslash}p{0.07\textwidth}
  >{\centering\arraybackslash}p{0.07\textwidth}
  >{\centering\arraybackslash}p{0.07\textwidth}
  >{\centering\arraybackslash}p{0.07\textwidth}
  >{\centering\arraybackslash}p{0.07\textwidth}
  >{\centering\arraybackslash}p{0.09\textwidth}
  >{\centering\arraybackslash}p{0.07\textwidth}
  >{\centering\arraybackslash}p{0.07\textwidth}
  >{\centering\arraybackslash}p{0.13\textwidth}
}
\toprule
Metrics & Methods & WX200 & Panda & UR5e & Bolt & Solo & PhantomX & Allegro & OP3 & \textbf{Mean}~$\boldsymbol{\pm}$~\textbf{Std} \\
\hline
\multirow{2}{*}{$\mathit{CD}\downarrow$} 
&Reart\cite{liu2023building} & 9.33 & 18.81 & 15.86 & 10.39 & 11.14 & 14.73 & 6.38 & 44.95 & 16.45 $\pm$ 12.18\\
&\textbf{Ours} & \textbf{7.49} & \textbf{13.56} & \textbf{12.84} & \textbf{8.41} & \textbf{9.77} & \textbf{10.88} & \textbf{5.80} & \textbf{8.30} & \textbf{9.63 $\pm$ 2.67} \\
\hline
\multirow{3}{*}{$\mathit{TED}\downarrow$} 
& MBS\cite{huang2021multibodysync} & 3.33 & 5.00 & 3.40 & 3.80 & 4.40 &  14.60 & 8.60 & 10.00 & 6.64 $\pm$ 4.07 \\
& Reart\cite{liu2023building} & 0.83 & 2.40 & 4.40 & 3.20 & 4.00 & 13.00 & 6.00 & 11.60 & 5.68$\pm$ 4.37 \\
& \textbf{Ours} & \textbf{0.33} & \textbf{1.40} & \textbf{0.60} & \textbf{1.75} & \textbf{0.00} & \textbf{4.00} & \textbf{4.00} & \textbf{6.00} & \textbf{2.26 $\pm$ 2.16} \\
\bottomrule
\end{tabular}}
\vspace{-0.5em}
\caption{Baseline Comparison.}
\vspace{-0.5em}
\label{tab:baseline_comparison}
\end{table*}

\begin{table*}[!t]
\centering
\captionsetup{width=\textwidth}
\resizebox{\textwidth}{!}{
\renewcommand{\arraystretch}{1.2}
\begin{tabular}{
  p{0.06\textwidth}|
  p{0.11\textwidth}|
  >{\centering\arraybackslash}p{0.07\textwidth}
  >{\centering\arraybackslash}p{0.07\textwidth}
  >{\centering\arraybackslash}p{0.07\textwidth}
  >{\centering\arraybackslash}p{0.07\textwidth}
  >{\centering\arraybackslash}p{0.07\textwidth}
  >{\centering\arraybackslash}p{0.09\textwidth}
  >{\centering\arraybackslash}p{0.07\textwidth}
  >{\centering\arraybackslash}p{0.07\textwidth}
  >{\centering\arraybackslash}p{0.13\textwidth}
}
\toprule
Metrics & Methods & WX200 & Panda & UR5e & Bolt & Solo & PhantomX & Allegro & OP3 & \textbf{Mean}~$\boldsymbol{\pm}$~\textbf{Std} \\
\hline
\multirow{3}{*}{$\mathit{CD}\downarrow$} 
& w/o anchor m & 7.95 & 14.67 & 14.19 & 9.74 & 10.73 & 12.31 & 6.38 & 9.86 & 10.73 $\pm$ 2.89\\
& w/o step m & \textbf{7.52} & 13.59 & 12.90 & 8.42 & 9.80 & 11.03 & \textbf{5.79} & 8.42 & 9.68 $\pm$ 2.69\\
& \textbf{Ours} & 7.55 & \textbf{13.52} &\textbf{12.84} & \textbf{8.39} & \textbf{9.78} & \textbf{10.93} &\textbf{5.79} & \textbf{8.32} & \textbf{9.64 $\pm$ 2.69}\\
\hline
\multirow{3}{*}{$\mathit{TED}\downarrow$} 
& w/o ori & 3.17 & 3.20 & 2.20 & 3.00 & 7.20 & 20.60 & 8.80 & 9.80 & 7.25 $\pm$ 6.14\\ 
& w/o pos & 1.33 & \textbf{0.60} & 0.80 & \textbf{1.40} & 6.20 & 17.40 & 7.60 & 12.00 & 5.92 $\pm$ 6.18\\ 
& \textbf{Ours} & \textbf{0.33} & 1.40 & \textbf{0.60} & 1.75 & \textbf{0.00} & \textbf{4.00} & \textbf{4.00} & \textbf{6.00} & \textbf{2.26 $\pm$ 2.16}  \\ 
\bottomrule
\end{tabular}}
\vspace{-0.5em}
\caption{Ablation Experiment.}
\vspace{-0.5em}
\label{tab:ab1}
\end{table*}

\begin{table*}[!t]
\centering
\captionsetup{width=\textwidth}
\resizebox{\textwidth}{!}{
\renewcommand{\arraystretch}{1.2}
\begin{tabular}{
  p{0.06\textwidth}|
  p{0.11\textwidth}|
  >{\centering\arraybackslash}p{0.07\textwidth}
  >{\centering\arraybackslash}p{0.07\textwidth}
  >{\centering\arraybackslash}p{0.07\textwidth}
  >{\centering\arraybackslash}p{0.07\textwidth}
  >{\centering\arraybackslash}p{0.07\textwidth}
  >{\centering\arraybackslash}p{0.09\textwidth}
  >{\centering\arraybackslash}p{0.07\textwidth}
  >{\centering\arraybackslash}p{0.07\textwidth}
  >{\centering\arraybackslash}p{0.13\textwidth}
}
\toprule
Metrics & Methods & WX200 & Panda & UR5e & Bolt & Solo & PhantomX & Allegro & OP3 & \textbf{Mean}~$\boldsymbol{\pm}$~\textbf{Std}
 \\
\hline
\multirow{2}{*}{$\mathit{CD}_r \downarrow$}
&1 Sequence & 11.51 & \textbf{42.82} & 35.76 & 17.86 & 13.41 & 14.76 & 8.66 & 32.25 & 22.13 $\pm$ 12.87 \\
&5 Sequences & \textbf{11.10} & 45.60 & \textbf{21.24} & \textbf{12.39} & \textbf{13.03} & \textbf{13.85} & \textbf{8.30} &\textbf{21.85} &\textbf{18.42 $\pm$ 11.96} \\
\hline
\multirow{2}{*}{$\mathit{E}_{JD} \downarrow$}
&1 Sequence & \textbf{1.16} & 4.46 & 3.32 & 5.44 & 5.34 & 4.39 & 6.20 & 11.60 & 5.24 $\pm$ 3.00\\
&5 Sequences & 1.17 & \textbf{3.95} & \textbf{2.17} & \textbf{4.94} & \textbf{1.61} & \textbf{2.50} & \textbf{3.97} & \textbf{7.37} & \textbf{3.46 $\pm$ 2.04} \\
\hline
\multirow{2}{*}{$\mathit{E}_{JA} \downarrow$}
&1 Sequence & 1.91 & 7.37 & 7.39 & 4.13 & 3.59 & 3.91 & 7.85 & 15.81 & 6.50 $\pm$ 4.34\\
&5 Sequences & \textbf{1.13} & \textbf{4.16} & \textbf{2.88} & \textbf{2.44} & \textbf{1.86} & \textbf{1.74}& \textbf{3.68} & \textbf{7.11} & \textbf{3.13 $\pm$ 1.90}\\
\bottomrule
\end{tabular}
}
\vspace{-0.5em}
\caption{Multi-Sequence Merging Experiment.}
\vspace{-0.5em}
\label{tab:seq}
\end{table*}

Figure \ref{fig_6} illustrates the core stages of AutoURDF, including the MST, segmentation results represented as cluster skeletons, joint prediction results, mesh quality, and simulation examples in the PyBullet simulator.
In Figure \ref{fig_7}, we qualitatively compare our segmentation and registration results with \textit{NPP} \cite{hayden2020nonparametric}, \textit{MultibodySync} (\textit{MBS})\cite{huang2021multibodysync}, and \textit{Reart} \cite{liu2023building} using the same point cloud sequences. Our method produces more accurate segmentation and matching results across all three robot sequences.
Figure \ref{fig_8} provides a direct comparison between the generated URDFs and the ground truth URDFs under the same new motor configurations. We also demonstrate the Silhouette Score with respect to DoF in Figure \ref{fig_9}, based on the Motion Correlation matrices. The highest score indicates the optimal number of distinct moving parts, and therefore the number of degrees of freedom.
Finally, we tested our method on real scanned point cloud data of a WX200 robot arm, commanding on all 5 servo motors, shown in Figure \ref{fig_10}. Despite the high noise in the scan and misalignment of the coordinate systems across frames, our method successfully derives a functional URDF file for this robot with only 10 frames data, achieving relatively accurate kinematics and morphology computation.

\subsection{Quantitative Results and Ablation Study}
\label{subsect:e3}
Table \ref{tab:baseline_comparison} shows a quantitative comparison between our method and two previous works \cite{huang2021multibodysync, liu2023building} on our validation dataset.
We compare our method with Reart for point cloud registration, using average end-point error measured in L1 Chamfer distance ($CD$). For robot body topology inference, we compare our method with both \textit{Reart} \cite{liu2023building} and \textit{MultiBodySync} (\textit{MBS}) \cite{huang2021multibodysync} by computing the tree editing distance ($TED$) between the ground truth kinematic tree and the predicted ones. Our method consistently achieves lower scores on both metrics, indicating improved registration accuracy and body topology inference accuracy across all types of robots.

To validate the contributions of the Step Model and Anchor Model in our point cluster registration algorithm, we compare the matching results of our complete pipeline with results using only the Step Model and only the Anchor Model. The complete method achieves superior performance, as shown in Table \ref{tab:ab1}.
Additionally, to validate the contributions of the two distance terms in Equation \ref{eq3}, we conduct ablation experiments comparing the complete equation with versions that include only the position term or only the orientation term. As shown in Figure \ref{fig_11}, using only the position distance struggles to produce accurate segmentation, while using only the orientation term is easily misled by links with similar rotational motions. Multiple links are segmented into the same group due to their similar rotational motion within this point cloud sequence.

Table \ref{tab:seq} presents a comparison of our method, extending the input point cloud data from 1 sequence (10 frames) to 5 sequences (50 frames). We evaluate the reposed robot shape error using the $CD_r$ metric, the joint axis normal distance with the $E_{JD}$ metric, and the joint angle error with the $E_{JA}$ metric. As the dataset scales, AutoURDF improves performance in 7 out of 8 robots for repose and joint distance and in all robots for joint angle accuracy. With 50 frames of point cloud data from the WX200 robot arm, we achieve an average joint angle difference of 1.13 degrees and a joint normal distance of 1.16 mm across its 5 joint axes.

Among all these improvements, AutoURDF stands out for its speed as an unsupervised method. On the WX200 robot data, registration takes \textbf{50} seconds, and URDF construction takes \textbf{12} seconds on an NVIDIA 3090 GPU. In comparison, \textit{Reart} \cite{liu2023building}  relax model and projection model training, with the flow model pre-trained, takes 35.5 minutes (2,120 seconds) on the same machine.

\section{Discussion and Conclusion}
\label{sec:conclusion}
{\bf{Limitations and future work:}}
Our method has three key limitations. First, we collected collision-free robot motion sequences, there is no self-collision or interaction with the environment, and thus our method does not learn dynamic information, and the resulting URDF files lack mass and moment of inertia data.
Second, segmenting more complex structures requires longer point cloud sequences. For example, with the OP3 humanoid robot, our method struggles to produce clean segmentation due to structural complexity. Extending the length of the point cloud video could improve segmentation accuracy for such complex robots.
Third, our method is limited to robots with revolute joints and tree-like kinematic structures. Future work could extend the approach to handle non-revolute joints—such as prismatic or spherical joints—as well as robots with parallel or closed-loop kinematic chains.\\

\noindent{\bf{Conclusion:}}
In this paper, we present a novel approach for constructing robot description files directly from point cloud data. 
Our method employs 6-DoF cluster registration to segment moving parts, infer body topology, and estimate joint parameters, achieving accurate joint direction estimation and whole-body shape reposing. 
The output URDF-formatted robot description file is compatible with widely used robot simulators. Experimental results show that our method outperforms previous approaches in the accuracy of robot point cloud registration and body topology estimation, offering an efficient and scalable solution for automated robot modeling from visual data.

\small
\noindent{\bf{Acknowledgements:}}
This work was supported in part by the US National Science
Foundation AI Institute for Dynamical Systems (DynamicsAI.org)
(grant no. 2112085). The authors thank Prof. Changxi Zheng and Ruoshi Liu for their invaluable feedback.

{
    \small
    \bibliographystyle{ieeenat_fullname}
    \bibliography{main}
}


\end{document}